\documentclass{article}
\DeclareUnicodeCharacter{2212}{-}
\usepackage[english]{babel}

\usepackage[letterpaper,top=2.54cm,bottom=2.54cm,left=2.54cm,right=2.54cm,marginparwidth=1.75cm]{geometry}

\usepackage{bm}
\usepackage{amsmath}
\usepackage{graphicx}
\usepackage{csquotes}
\usepackage{tabularx}
\usepackage{hanging}
\usepackage[colorlinks=true, allcolors=blue]{hyperref}

\usepackage{authblk}
\title{Can Moran Eigenvectors Improve Machine Learning of Spatial Data? Insights from Synthetic Data Validation}
\author[1,2]{Ziqi Li}
\author[3]{Zhan Peng}
\affil[1]{Department of Geography, Florida State University, United States}
\affil[2]{The Spatial Data Science Center, Florida State University, United States}
\affil[3]{Transportation and Geographic Information Science Lab, Graduate School of Information Sciences, Tohoku University, Japan}

\begin{document}
\maketitle

\noindent Citation: Li, Z. and Peng, Z. (forthcoming). Can Moran Eigenvectors Improve Machine Learning of Spatial Data? Insights from Synthetic Data Validation. \textit{Geographical Analysis}.

\begin{abstract}
Moran Eigenvector Spatial Filtering (ESF) approaches have shown promise in accounting for spatial effects in statistical models. Can this extend to machine learning? This paper examines the effectiveness of using Moran Eigenvectors as additional spatial features in machine learning models. We generate synthetic datasets with known processes involving spatially varying and nonlinear effects across two different geometries. Moran Eigenvectors calculated from different spatial weights matrices, with and without a priori eigenvector selection, are tested. We assess the performance of popular machine learning models, including Random Forests, LightGBM, XGBoost, and TabNet, and benchmark their accuracies in terms of cross-validated R² values against models that use only coordinates as features. We also extract coefficients and functions from the models using GeoShapley and compare them with the true processes. Results show that machine learning models using only location coordinates achieve better accuracies than eigenvector-based approaches across various experiments and datasets. Furthermore, we discuss that while these findings are relevant for spatial processes that exhibit positive spatial autocorrelation, they do not necessarily apply when modeling network autocorrelation and cases with negative spatial autocorrelation, where Moran Eigenvectors would still be useful.

\end{abstract}

\section{Introduction}

Spatial data are special because of the way they are distributed in space and how they are generated by underlying processes. Modeling spatial data requires additional attention due to two well-recognized spatial effects: spatial autocorrelation and spatial heterogeneity (Anselin, 1988). Ignoring these effects in standard statistical frameworks can lead to biased estimates and incorrect inferences. To address these issues, a wide array of spatial statistical models have been developed, including spatial econometric models (Anselin and Bera, 1998; LeSage and Pace, 2009), Geographically Weighted Regression (GWR) and Multi-scale GWR (Fotheringham et al., 2002; Fotheringham et al., 2017; Fotheringham et al., 2023), Kriging-based methods (Kammann and Wand, 2003; Hengl et al., 2007), Eigenvector Spatial Filtering (ESF) approaches (Tiefelsdorf and Griffith, 2007; Griffith and Chun, 2014; Seya et al., 2015; Murakami et al., 2017;  Peng and Inoue, 2024). Among these, ESF approaches show promise in accounting for spatial autocorrelation and heterogeneity and have been extensively applied in various fields such as ecology (Diniz‐Filho and Bini, 2005; Kühn et al., 2009), environmental modeling  (Tan et al., 2020; Shi et al., 2021), crime analysis (Helbich and Jokar Arsanjani, 2015, Chun, 2014), real estate (Helbich and Griffith, 2016; Peng and Inoue, 2022), among others. 

The essence of ESF is to use eigenvectors derived from a spatial weights matrix as additional independent variables in a linear regression model. The spatial weight matrix is intended to represent the spatial structure of the data generation process, and the resulting eigenvectors will exhibit different patterns of spatial autocorrelation. By incorporating these eigenvectors into a regression model, ESF can effectively filter out spatial autocorrelation, leading to more accurate and reliable statistical inference. Results from Griffith and Chun (2014) indicate that ESF produces outcomes comparable to those of spatial lag and error models. The ESF approach can also be applied to Spatially Varying Coefficient models (ESF-SVC) to account for spatial process heterogeneity. ESF-SVC includes interaction terms between Moran Eigenvectors and independent variables so an additive combination of Moran Eigenvectors will serve as the spatially varying coefficients associated with each independent variable (Griffith, 2008). Comparison studies in the literature show that ESF-SVC achieves results similar to those of GWR and MGWR (Oshan and Fotheringham, 2017; Murakami, 2019).

In addition to statistical framework, recent research explored the use of Moran Eigenvectors as additional features (i.e., independent variables) in machine learning models. Islam et al. (2022) added Moran Eigenvectors to XGBoost, Neural Network, and Support Vector Machine models and tested them on an empirical housing price dataset of King County. They concluded that the inclusion of Moran Eigenvectors improved the respective machine learning models' R² by 0.2 to 0.25, and substantially reduced spatial autocorrelation in the residuals from the machine learning models to a negligible level. Similarly, Liu et al. (2022) found that Random Forest models with Moran Eigenvectors yielded lower errors (up to 33\% difference) and reduced the global spatial autocorrelation of the residuals (up to a 95\% decrease in Moran’s I) compared to the RF model with only non-spatial features based on the meuse soil data and the California housing data. In addition, Hu et al. (2022) included Moran Eigenvectors in a Random Forest model to predict house prices in Fairfax, Virginia, and reported that the testing accuracy of the Random Forest model with eigenvectors achieved the best accuracy (R²=0.887) compared to the model with x-y coordinates of the property (R²=0.883) and models with only non-spatial features (R²=0.855). In summary, all three works demonstrate that adding Moran Eigenvectors is useful in machine learning models to account for spatial effects compared to models without them. However, neither Islam et al. (2022) nor Liu et al. (2022) compared the model with just x-y coordinates as spatial information against the eigenvector approach. The results from Hu et al. (2022), which include this comparison, showed only marginal benefits of using Moran Eigenvectors instead of coordinates. Considering that including only x-y coordinates of locations would be much simpler for machine learning models in terms of both computational efficiency and convergence, it may be worth further investigation into the effectiveness and utility of  Moran Eigenvectors in machine learning.

Another gap in previous research concerns the aspect of model explainability. Since machine learning models are often considered 'black boxes,' understanding how they handle spatial features and spatial effects can be challenging. Recent developments in the field of Explainable AI (XAI) have provided valuable tools to delve into machine learning models, among which SHapley Additive exPlanations (SHAP) have gained substantial attention (Lundberg and Lee, 2017). SHAP applies the Shapley value concept from game theory to machine learning models to quantify feature importance at both global (averaged across all observations) and local (at individual observation) levels. SHAP has seen an increasing number of applications in spatial machine learning (e.g., Pradhan et al., 2023; Li, 2023; Mete and Yomralioglu, 2023; Xue et al., 2024). Specifically, the local explanation nature of SHAP provides a way to measure how feature importance varies over space. In this regard, as a first attempt, Li (2022) using synthetic examples with known data-generating processes find that SHAP method can be used to extract spatial effects and that the coefficients derived from the XGBoost model are comparable to those from statistical models such as the spatial lag model and MGWR. However, the off-the-shelf SHAP method is not spatially explicit, as it does not consider spatial features to be different from non-spatial features. Another limitation is that its current implementation does not support extracting spatial effects for non-tree-based models (e.g., Neural Networks). To address this, Li (2024) further developed GeoShapley, which extends the SHAP framework together with Joint Shapley value and Shapley interaction value, to provide a truly model-agnostic tool that can measure spatial effects from any machine learning model. GeoShapley is able to quantify the importance of a collective set of spatial features, whether they are location coordinates, a set of Moran Eigenvectors, or more complicated spatial embedding methods (e.g., as discussed in Mai et al., 2022). The resulting GeoShapley values have interpretations that are directly linked with SVC models for explaining spatial effects and additive models for explaining non-spatial effects. In this regard, GeoShapley can serve as a diagnostic tool to cross-compare and evaluate machine learning models at the coefficient and function levels instead of merely regarding overall model accuracy, which can facilitate understanding the effectiveness of Moran Eigenvectors in machine learning models.

Accordingly, the overarching aim of this paper is to evaluate whether integrating Moran Eigenvectors into machine learning models improves their ability to account for spatial effects and enhance accuracy. We compare models using simple coordinates as the baseline and models that incorporate various specifications of Moran Eigenvectors. We select state-of-the-art machine learning models for tabular regression tasks, including decision tree-ensemble methods such as Random Forest, XGBoost, LightGBM, and the deep-learning method TabNet. We train these models on synthetic datasets with known generating processes and explain the models' outputs using GeoShapley. We compare the accuracy of the overall model performance and visualize how well each model replicates the generating processes. The comparisons with different datasets and models provide a holistic view of the effectiveness of Moran Eigenvectors in the context of machine learning.

The remainder of this paper is structured as follows: Section 2 provides the basics on Moran Eigenvectors. Section 3 describes the details of data and methods used. Section 4 presents the results and their analysis. Finally, Section 5 discusses the findings and concludes the paper.

\section{Moran Eigenvectors Basics}

Moran Eigenvectors are fundamental components that represent the entire spectrum of spatial relationships of the data. Following Griffith (2003), Moran Eigenvectors are calculated through the following eigen-decomposition of a centered spatial weights matrix:
\begin{equation}\label{eq:MCM}
    (\mathbf{I}-\mathbf{1}\mathbf{1}^\prime/n)\mathbf{C}(\mathbf{I}-\mathbf{1}\mathbf{1}^\prime/n)=\mathbf{E}_n\mathbf{\Lambda}\mathbf{E}_n^\prime
\end{equation}
The spatial weights matrix \(\mathbf{C}\) is an $n\times n$ symmetric matrix which reflects the spatial structure of data observed at \(n\) locations. Each entry $c_{ij}$ is a spatial weight that defines the spatial relation between each pair of locations $(i, j)$. $c_{ij}$ is mainly calculated based on contiguity or distance between two locations. The simplest way of the first case is to define $c_{ij}=1$ if two neighbor locations share an edge, otherwise $c_{ij}=0$. In the second case, the value of $c_{ij}$ is 1 if two neighbor locations are within a fixed range, or decreases as the distance between two neighbor locations increases, according to a specific kernel-weighted function (e.g., exponential, Gaussian). The $n\times n$ matrix  $\mathbf{E}_n$ includes all Moran Eigenvectors $[\mathbf{e}_1,\ldots,\mathbf{e}_n]$, and the $n\times n$ diagonal matrix $\mathbf{\Lambda}$ holds the corresponding eigenvalues $\left\{\lambda_1,\ldots,\lambda_n\right\}$. Additionally, $\mathbf{I}$ denotes an $n\times n$ identity matrix, and $\mathbf{1}$ is an $n\times 1$ vector of ones. The spatial weights matrix \(\mathbf{C}\) is centered by \((\mathbf{I}-\mathbf{1}\mathbf{1}^\prime/n)\) to ensure that the resulting Moran Eigenvectors are orthogonal and uncorrelated. Therefore, each eigenvector represents a distinct pattern of spatial autocorrelation, with the scale proportional to the corresponding eigenvalue. Specifically, the first Moran eigenvector with the largest positive eigenvalue exhibits the most global scale, while the scale becomes more localized as the eigenvalue decreases.

Using non-spatial approaches to model spatial data leads to inefficient estimates due to the underlying spatial autocorrelation in the model's error term. To solve this problem, Griffith (2003) proposed the ESF model by incorporating Moran Eigenvectors as additional variables in a regression model. According to the aforementioned characteristics, these Moran Eigenvectors function as spatial filters to eliminate spatial autocorrelation in the error, resulting in only random noise, which helps to mitigate potential estimation bias. The ESF model is formulated as:
\begin{equation}\label{eq:esf}
\mathbf{y}=\mathbf{X}\bm{\beta}+\mathbf{E}\bm{\gamma}+\bm{\varepsilon},\ \bm{\varepsilon} \sim \mathcal N(\mathbf{0},\sigma^2\mathbf{I})
\end{equation}

where $\mathbf{y} = [{y}_1,\ldots,{y}_n]^\prime$ is an $n\times 1$ vector of dependent variable, $\mathbf{X}=[\mathbf{X}_0,\mathbf{X}_1,...,\mathbf{X}_K]$ is an $n\times (K+1)$ matrix of independent variables ($\mathbf{X}_0$ is a vector of ones for the intercept). The matrix $\mathbf{E}$ is an $n\times L$ matrix of a subset of  $\mathbf{E}_n$, including only $L$ eigenvectors $[\mathbf{e}_1,...,\mathbf{e}_L]$ with positive eigenvalues, as positive spatial autocorrelation is more common in most spatial analyses. Note that $L<n$.  $\bm{\beta}=[\beta_{0}\ldots,\beta_{K}]^\prime$ and $\bm{\gamma}=[\gamma_{1}\ldots,\gamma_{L}]^\prime$ are $(K+1)\times 1$ and $L\times 1$ vectors of regression coefficients, respectively. $\bm{\varepsilon}$ is the random error following the normal distribution. The estimation of ESF can be easily done by using ordinary least squares (OLS) approach.

Furthermore, to consider spatial heterogeneity, Griffith (2008) extended the above ESF model into the ESF-SVC model (Equation (\ref{eq:esfsvc})) by interacting Moran Eigenvectors with each independent variable. 
\begin{equation}\label{eq:esfsvc}
    \mathbf{y}=\sum_{k\ =\ 0}^{K}{\mathbf{X}_k\circ\left(\beta_k\mathbf{1}\ +\ \mathbf{E}\bm{\gamma}_k\right) +\bm{\varepsilon}},\ \bm{\varepsilon} \sim \mathcal N(\mathbf{0},\sigma^2\mathbf{I})
\end{equation}

Here, ``$\circ$'' denotes the element-wise product operator. For the coefficient of each independent variable $\mathbf{X}_k$, \(\beta_k\mathbf{1}\) represents an average trend that is uniform for all observations in space, and the linear combination of Moran Eigenvectors \(\mathbf{E}\bm{\gamma}_k\) accounts for spatial heterogeneity, which is also the local deviation from the global trend. Parameters $\beta_k$ and $\bm{\gamma}_k=[\gamma_{k1}\ldots,\gamma_{kL}]^\prime$ are estimated for each $\mathbf{X}_k$ to capture the unique pattern of each spatially varying coefficient.

Notably, the eigen-decomposition shown in Equation (\ref{eq:MCM}) is slow when the data set is large. Hence, in this study, we adopt an approach proposed by Murakami and Griffith (2020) to accelerate the calculation. This approach extracts a smaller set of Moran Eigenvectors from the complete sample locations based on Nystrom extension, thereby greatly minimizing the computation time.  A simulation study demonstrated that 200  eigenvectors are sufficient to represent the most positively correlated variations, even for large samples such as \(n\) = 100,000 (see appendix of Murakami and Griffith, 2019a). Furthermore, due to the interaction term in the ESF-SVC model, incorporating all eigenvectors will result in a substantial rise in the number of variables in the model, thereby heightening the risk of overfitting. Additionally, not every Moran eigenvector is essential for explaining the spatial heterogeneity of each coefficient. For instance, if a coefficient's value varies at a global spatial scale, eigenvectors with large eigenvalues will be dominant in modeling this pattern, while those with small eigenvalues may not have significant effects. Therefore, it is recommended to select a subset of eigenvectors for each coefficient to represent its distinct spatial variation while maintaining model simplicity. This subset is typically specified via the forward stepwise selection (Griffith and Chun, 2014) or the least absolute shrinkage and selection operator (LASSO) (Seya et al., 2015, Peng and Inoue, 2022). Both methods aim to eliminate unnecessary eigenvectors from the model. However, LASSO achieves this by shrinking specific coefficients $\bm{\gamma}_k$ to zero and is considered more efficient compared to the stepwise procedure (Seya et al., 2015).

\section{Data and Methods}

\begin{figure}[h]
    \centering
    \includegraphics[width=1\linewidth]{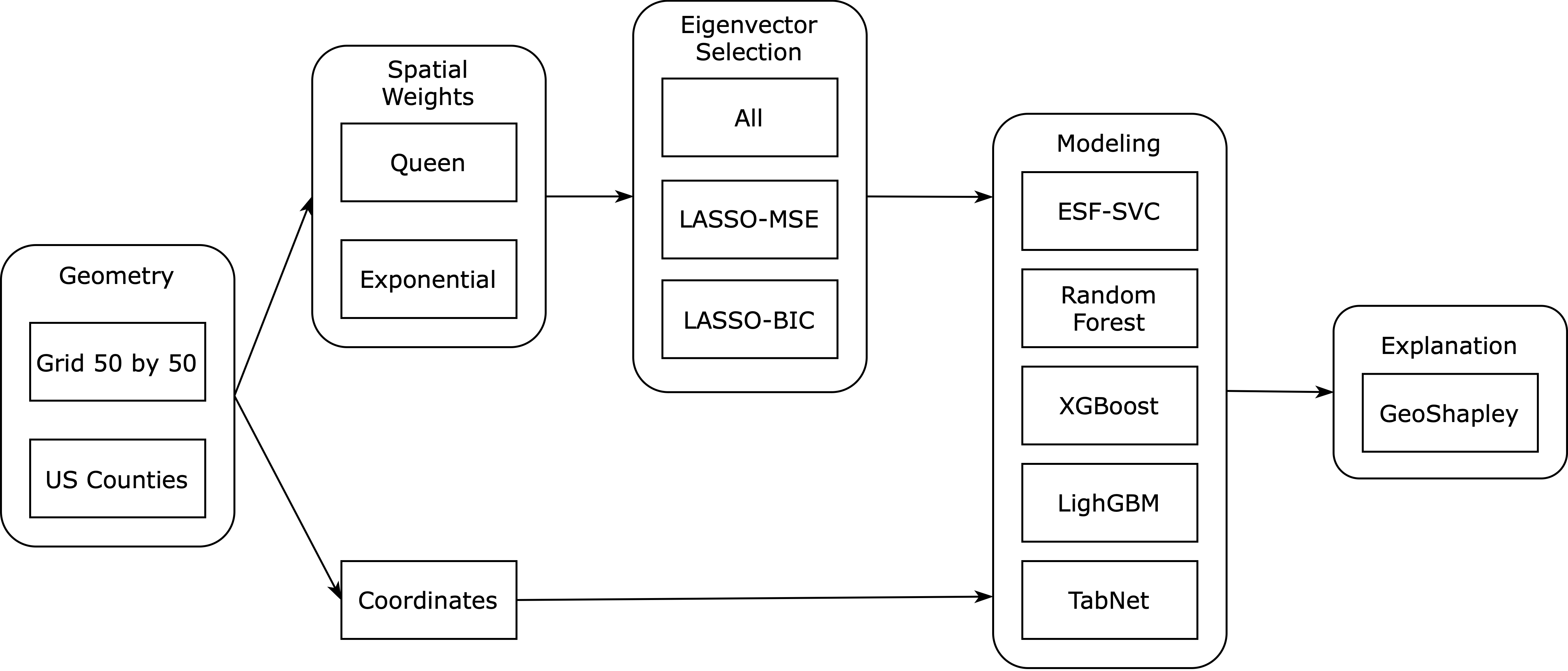}
    \caption{Workflow}
    \label{fig:workflow}
\end{figure}

We generate synthetic data based on known data-generating processes to benchmark multiple scenarios explored in this paper. Specifically, we design two separate geometries, using spatial features from location coordinates or from Moran Eigenvectors based on different spatial weights matrices, with different eigenvector selection approaches and test with different modeling frameworks. All models will then be compared regarding their predictive accuracy and explained by GeoShapley to allow a closer examination at the process level. A detailed workflow is presented as  \ref{fig:workflow}

\subsection{Data generating processes}

Consider that in the real world, spatial data are often as the result of a mixture of complex spatial and nonlinear processes. Accordingly, we design the data generating processes as follows:

\begin{equation}\label{eq:dgp}
\mathbf{y} = 3 + (\bm{\beta}_1\circ \mathbf{X}_1 + \mathbf{X}_1^2) + (\bm{\beta}_2\circ  \mathbf{X}_2 + 2\mathbf{X}_2) + \bm{\varepsilon}
\end{equation}

where the dependent variable $\mathbf{y}$ is a function of a global intercept value of 3, two features $\mathbf{X}_1$ and $\mathbf{X}_2$, and a random error $\mathbf{\varepsilon}$. Specifically, the association between feature $\mathbf{X}_1$ and the dependent variable includes a location-specific spatial effect $\bm{\beta}_1$, and a location-invariant global non-linear effect. Similarly, the association between feature $\mathbf{X}_2$ and the dependent variable includes a location-specific spatial effect $\bm{\beta}_2$, and a location-invariant global linear effect. The two location-specific spatial effects, $\bm{\beta}_1$ and $\bm{\beta}_2$, are generated by Gaussian Random Fields $GRF(0, \mathbf{\Omega})$ with a mean of zero and a covariance matrix $\mathbf{\Omega}$, defined as

\begin{equation}\label{eq:grf}
\mathbf{\Omega}(l) = \exp\left( -0.5 \left( \frac{\mathbf{d}}{l} \right)^2 \right)
\end{equation}

where $\mathbf{d}$ is an $n \times n$ matrix containing pairwise distances for all locations, and $l$ is a scale parameter indicating the amount of distance-decay in the covariance function. Process $\bm{\beta}_1$ is generated with $l=8$, which exhibits local variation, and process $\bm{\beta}_2$ is simulated with $l=12$, yielding a more long-range regional spatial effect. $\mathbf{X}_1$  and $\mathbf{X}_2$  are drawn from a uniform distribution $U(−2,2)$, and the error term is drawn from a normal distribution $N(0,0.5)$. The dependent variable $\mathbf{y}$ can then be reconstructed using the features and the designed processes.

\begin{figure}
\centering
\includegraphics[width=1\linewidth]{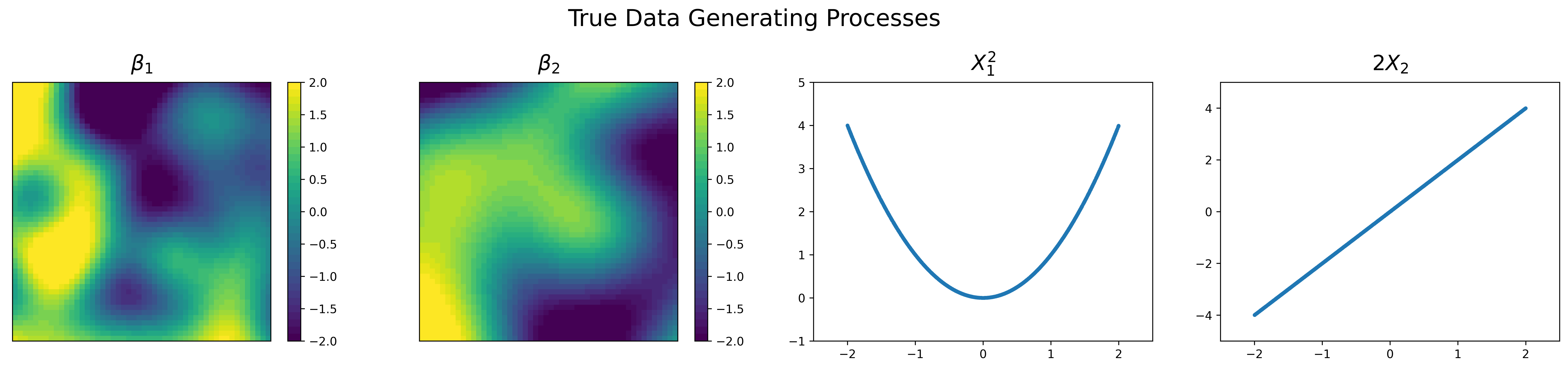}
\caption{\label{fig:grid_true}True data generating processes for Grid.}
\end{figure}

\begin{figure}
\centering
\includegraphics[width=1\linewidth]{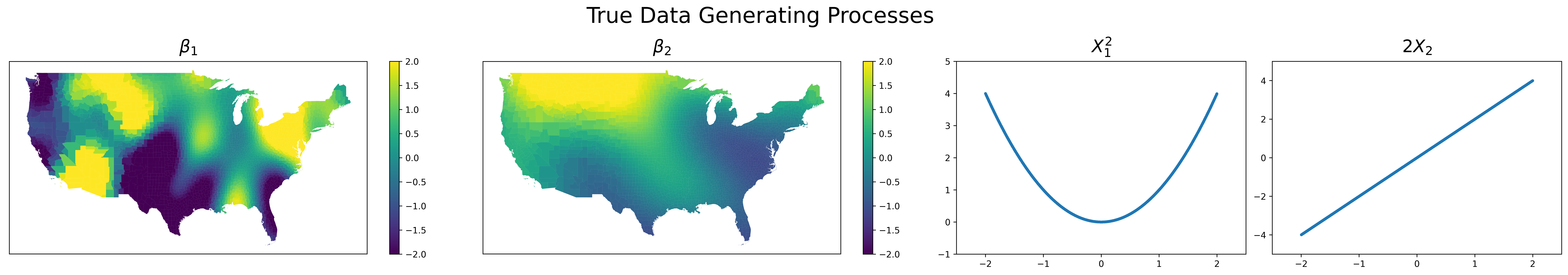}
\caption{\label{fig:us_true}True data generating processes for US Counties .}
\end{figure}

We also generate data for two separate geometries common in spatial data: 1) a regularly sampled grid dataset with a size of 50 by 50, and 2) a real-life irregularly sampled polygon dataset using the 3,109 counties in the contiguous United States. We refer to the first dataset as \textit{"Grid"} and the second dataset as \textit{“US Counties”} throughout the paper. The two datasets and their DGPs can be seen in \ref{fig:grid_true}  and  \ref{fig:us_true}

\subsection{Moran eigenvectors and selection}
\begin{figure}
    \centering
    \includegraphics[width=0.75\linewidth]{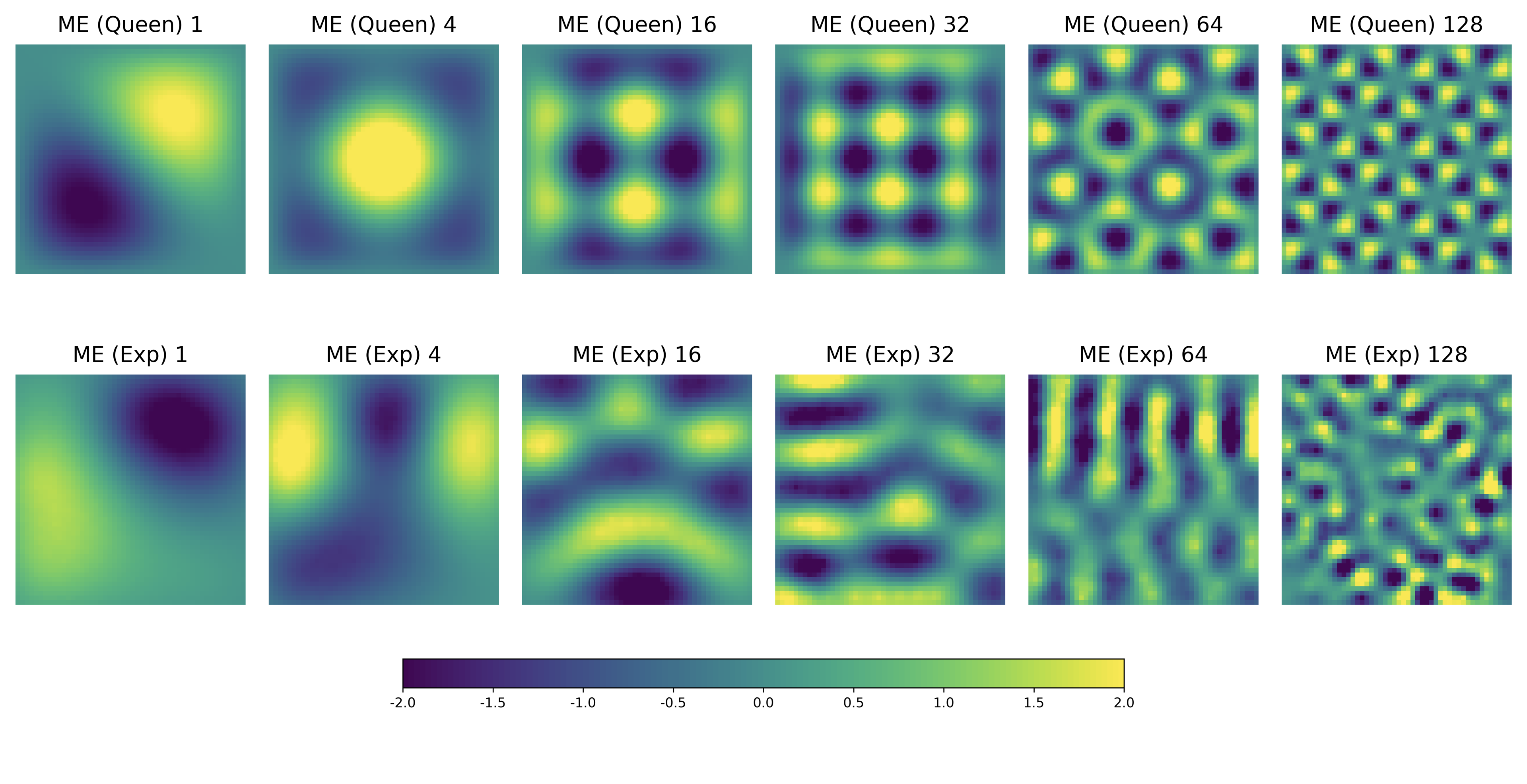}
    \caption{Illustration of selected Moran Eigenvectors for Grid}
    \label{fig:grid_mem}
\end{figure}

\begin{figure}
    \centering
    \includegraphics[width=0.75\linewidth]{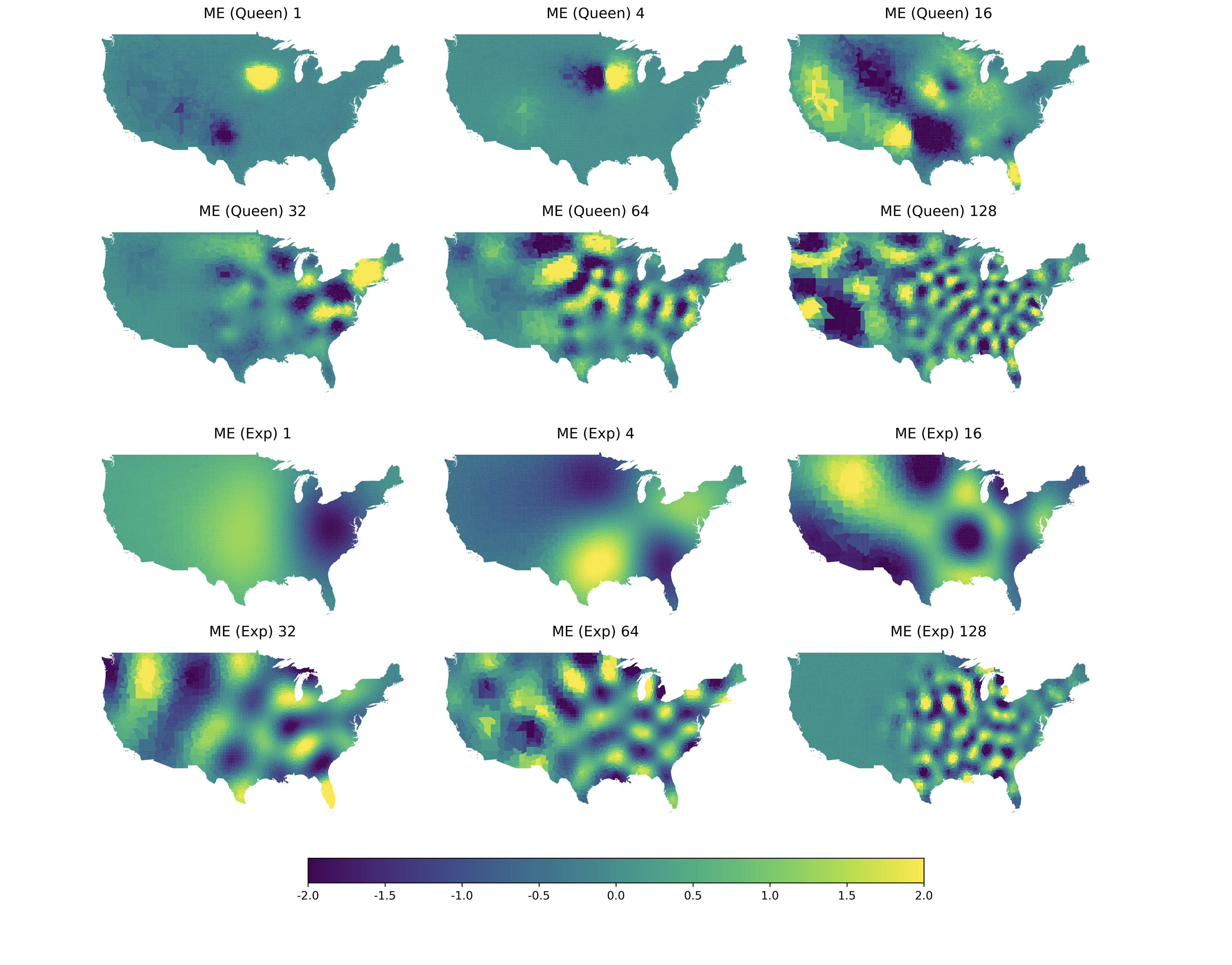}
    \caption{Illustration of selected Moran Eigenvectors for US Counties}
    \label{fig:us_mem}
\end{figure}

The specification of the spatial weights matrix will result in different patterns of Moran Eigenvectors. Here, we calculate Moran Eigenvectors based on two common spatial weights matrices: 1) a Queen spatial contiguity-based matrix and 2) an exponential distance kernel-based matrix. Following Murakami et al. (2017), this matrix is defined by $c_{i,j}=\exp{\left(-d_{i,j}/{r}\right)}$, where $d_{i,j}$ is the Euclidean distance between locations $i$ and $j$, and $r$ as the longest distance in the minimum spanning tree connecting all locations. The Queen spatial weights matrix and Moran Eigenvectors are calculated using the \textit{PySAL} Python package (Rey et al., 2022), while the exponential distance kernel-based matrix is calculated using the \textit{spmoran} R package (Murakami, 2017). The top 200 Moran Eigenvectors are used as the candidate set. For illustration, the 1st, 4th, 16th, 32nd, 64th and 128th Moran Eigenvectors are visualized in \ref{fig:grid_mem} and \ref{fig:us_mem}.

We use LASSO to select a subset of Moran Eigenvectors that best represent the spatial processes underlying the data. We experimented with two LASSO optimization criteria: 1) 5-fold cross-validation to minimize the mean-squared error (LASSO-MSE) and 2) the Bayesian Information Criterion (LASSO-BIC). The LASSO modeling is conducted using the \textit{scikit-learn} package in Python. The subset of Moran Eigenvectors with non-zero coefficients from the best LASSO model will be used for subsequent analysis. We also include a scenario with no eigenvectors selection (i.e. using all 200 Moran Eigenvectors) as a benchmark.

\subsection{Machine learning models}
We include four popular and performance-leading machine learning models in the comparison: Random Forest (Breiman, 2001), XGBoost (Chen and Guestrin, 2016), LightGBM (Ke et al., 2017), and TabNet (Arik and Pfister, 2021). Random Forest is based on bootstrap aggregating (bagging) decision trees with random sampling of training data, which are then ensembled through an averaging approach. XGBoost is an optimized implementation of gradient boosting trees that builds decision trees sequentially to correct errors made by previously trained trees. LightGBM, developed by Microsoft, is another high-performance gradient boosting framework that uses a histogram-based approach to find optimal splits, making it faster and more efficient with large datasets. The three tree-based models are selected due to their superior performance in handling tabular data tasks and their reported use cases in handling spatial data (e.g., Hengl et al., 2018; Zhong et al., 2021; Ma et al., 2021). To complement the perspective from a non-ensemble tree-based approach, we include TabNet, a deep learning model for tabular data developed by Google, which leverages sequential attention mechanisms to selectively choose features at each decision step. It is similar to decision trees in the feature selection and decision steps but within a neural network framework. It excels in handling a variety of tabular data problems and achieves state-of-the-art performance on a range of datasets. In this way, the comparison offers a more holistic view of the compatibility of Moran Eigenvectors with two different machine learning architectures. 

Two baseline models are also used, including the linear regression model with no spatial features (using only features $X_1$ and $X_2$), and ESF-SVC. Random Forest, LightGBM, and XGBoost are trained using Microsoft’s \textit{FLAML} Python library, which is a lightweight solution for efficient and automatic machine learning model tuning (Wang et al., 2021). TabNet is trained using the \textit{pytorch-tabnet} Python library. For all models, a 5-fold cross-validation approach is used for hyperparameter tuning, and the cross-validated out-of-sample $R^2$ is reported for overall model accuracy assessment. All the models are trained on Google Colaboratory with its High-RAM kernel. For a fair comparison, a time budget of 20 minutes is applied to each model training session.

\subsection{GeoShapley}
ESF-SVC and machine learning models are then explained by GeoShapley. GeoShapley is a post-hoc explanation method that measures spatial and non-spatial effects from any machine learning model. GeoShapley is based on the Shapley value, which originates from cooperative game theory to quantify players’ contributions toward a cooperative game. The core concept in GeoShapley is that spatial features, whether they are coordinates or a set of Moran Eigenvectors, are considered as a joint player along with other features in a model prediction game. Then, the importance of spatial features and their interactions with other features can be quantified, resulting in potential spatially varying effects. GeoShapley combines Shapley interaction values and Joint Shapley values (Harris et al., 2021) and is estimated using the Kernel SHAP estimator developed by Lundberg and Lee (2017). More technical details can be found in Li (2024). Essentially, GeoShapley breaks down model prediction $\mathbf{\hat{y}}$ into the following four additive components:

\begin{equation}\label{eq:geoshap}
\mathbf{\hat{y}} = \phi_0 + \mathbf{\phi_{GEO}} + \sum_{j=1}^{p} \mathbf{\phi_j} + \sum_{j=1}^{p} \mathbf{\phi_{(GEO,j)}}
\end{equation}

where 1) $\phi_0$ is a constant value analogous to the intercept in a linear regression model; 2) $\phi_{GEO}$ is a vector of size $n$ measuring the intrinsic location effect in the model; 3) $\mathbf{\phi_j}$ is a vector of size $n$ for each non-location feature $\mathbf{X_j}$, providing a location-invariant effect to the model, such as global linear or non-linear effects; and $\mathbf{\phi_{(GEO,j)}}$ is a vector of size $n$ for each non-location feature $\mathbf{X_j}$, providing the spatially varying interaction effect to the model. $\mathbf{\phi_{(GEO, j)}}$ does not directly yield spatially varying coefficients, but they can be obtained by a univariate GWR smoother to regress $\mathbf{\phi_{(GEO,j)}}$ against $\mathbf{X_j}$. GeoShapley values are calculated using the \textit{geoshapley} Python package (Li, 2024).

\subsection{Code and reproducibility}
All the code that generates the data, performs model training and explanation, and obtains the reported results and figures can be found at this repository: \href{https://anonymous.4open.science/r/Machine_learning_moran_eigenvector-01FF/README.md}{Code and Data Repository}

\section{Results}
\subsection{Model accuracy}

Table 1 and Table 2 show the comparison of model accuracy in various scenarios regarding the cross-validated out-of-sample R² for the Grid and US Counties datasets, respectively. The tables are organized with models using Moran Eigenvectors based on either Queen or Exponential kernel specifications of the spatial weights matrix, and with no variable selection, LASSO selection based on the MSE criterion, and LASSO selection based on the BIC criterion. The number in bracket for each LASSO based approach refers to the number of Moran Eigenvectors selected as features in the model by LASSO. "Coords" refers to models that use only coordinates as spatial features.

In Table 2, a linear model with no spatial input or effects yields an R² of 0.465. Across all methods, TabNet with coordinates as the only spatial features is the best-performing model with an R² of 0.955. Other observations can be made: 1) Machine learning models, including Random Forest, LightGBM, and XGBoost, generally outperform ESF-SVC due to the fact that the latter does not account for the non-linear relationships in the data generating process.  2) Among tree-based models, XGBoost and LightGBM perform similarly, with both models consistently outperforming Random Forests in all scenarios. For example, under the Queen-based weight matrix and LASSO-BIC selection, XGBoost and LightGBM achieve R² values of 0.911 and 0.909, respectively, while Random Forest yields an R² of 0.818. 3) Pre-selection of Moran Eigenvectors is beneficial for all machine learning models, with the BIC-based approach generally outperforming no selection or MSE-based approaches. This is particularly evident for TabNet, where the R² improves from 0.529 to 0.837 under the Queen-based weight matrix when switching from all variables to LASSO-BIC pre-selection. 4) Exponential kernel-based Moran Eigenvectors consistently leads to better model performance than Queen-based Moran Eigenvectors, as seen across all models. 5) Finally, models that use coordinates outperform models using Moran Eigenvectors. The results for the US Counties geometry which are shown in Table 3 generally mirror the results for the Grid geometry.

\begin{table}
\footnotesize
\centering
\caption{Model accuracy comparison for the Grid dataset based on cross-validated out-of-sample R².}
\label{tab:my_table}
\begin{tabular}{|>{\raggedright\arraybackslash}p{0.13\linewidth}|>{\raggedleft\arraybackslash}p{0.07\linewidth}|>{\raggedleft\arraybackslash}p{0.11\linewidth}|>{\raggedleft\arraybackslash}p{0.11\linewidth}|>{\raggedleft\arraybackslash}p{0.07\linewidth}|>{\raggedleft\arraybackslash}p{0.11\linewidth}|>{\raggedleft\arraybackslash}p{0.11\linewidth}|>{\raggedleft\arraybackslash}p{0.06\linewidth}|} \hline 

Grid & \multicolumn{3}{|c|}{Queen}& \multicolumn{3}{|c|}{Exp}&   \\ \hline 

 & All (200) & LASSO-MSE (172)& LASSO-BIC (64)& All  (200) & LASSO-MSE (154)& LASSO-BIC (60)& Coords \\ \hline 

ESF-SVC& 0.748& 0.795 & 0.772 & 0.760& 0.807& 0.791& NA \\ \hline 

Random Forest & 0.788 & 0.790 & 0.818 & 0.788& 0.796& 0.822& 0.860 \\ \hline 

LightGBM & 0.893 & 0.894 & 0.909 & 0.898& 0.895& 0.915& 0.942 \\ \hline 

XGBoost & 0.888& 0.892& 0.911& 0.888& 0.895& 0.914& 0.935\\ \hline 

TabNet & 0.529& 0.586& 0.837& 0.574& 0.661& 0.872& 0.955\\ \hline 

Linear & \multicolumn{7}{|r|}{0.465} \\ \hline

\end{tabular}

\end{table}

\begin{table}
\centering
\footnotesize
\caption{Model accuracy comparison for the US Counties dataset based on cross-validated out-of-sample R².}
\begin{tabular}{|>{\raggedright\arraybackslash}p{0.13\linewidth}|>{\raggedleft\arraybackslash}p{0.07\linewidth}|>{\raggedleft\arraybackslash}p{0.11\linewidth}|>{\raggedleft\arraybackslash}p{0.11\linewidth}|>{\raggedleft\arraybackslash}p{0.07\linewidth}|>{\raggedleft\arraybackslash}p{0.11\linewidth}|>{\raggedleft\arraybackslash}p{0.11\linewidth}|>{\raggedleft\arraybackslash}p{0.06\linewidth}|} \hline 

US Counties & \multicolumn{3}{|c|}{Queen} & \multicolumn{3}{|c|}{Exp} &   \\ \hline 

 & All (200) & LASSO-MSE(177)& LASSO-BIC (115)& All (200) & LASSO-MSE (149)& LASSO-BIC (92)& Coords \\ \hline 

ESF-SVC& 0.755& 0.791& 0.760 & 0.797& 0.832& 0.819& N/A \\ \hline 

Random Forest & 0.717& 0.721& 0.736& 0.772& 0.782& 0.796& 0.842 \\ \hline 

LightGBM & 0.848 & 0.854& 0.873& 0.871& 0.882& 0.894& 0.931\\ \hline 

XGBoost & 0.841 & 0.844& 0.862& 0.876& 0.873& 0.900& 0.929\\ \hline 

TabNet & 0.769& 0.798& 0.836& 0.769& 0.832& 0.889& 0.948\\ \hline 

Linear & \multicolumn{7}{|r|}{0.471} \\ \hline

\end{tabular}

\end{table}

\subsection{GeoShapley explanations}

\begin{figure}
    \centering
    \includegraphics[width=0.75\linewidth]{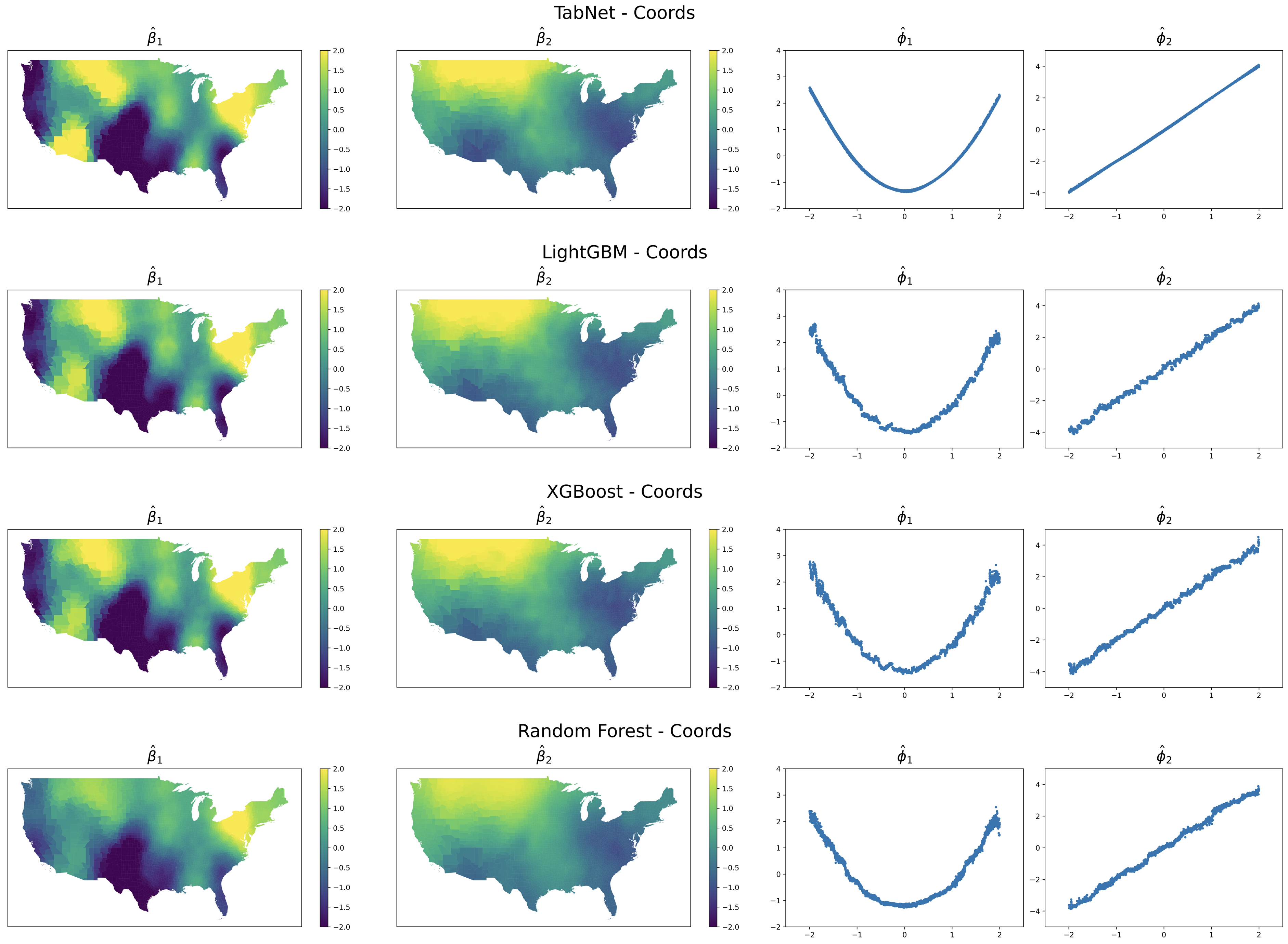}
    \caption{GeoShapley explanations for four models using coordinates as input spatial features.}
    \label{fig:geoshapley_us_coords}
\end{figure}

In addition to model-level accuracy comparisons, with GeoShapley, we are able to explain the spatially varying and global effects at the parameter level for each model. Figure \ref{fig:geoshapley_us_coords} presents model explanations to models using only coordinates for the US Counties dataset. The top performing models, TabNet, LightGBM, and XGBoost, accurately replicate the true data-generating process as originally shown in Figure \ref{fig:us_true} . For example, regarding $\beta_1$, machine learning models can capture the strong positive effects in states such as Arizona, Montana, the Dakotas, and states in the Northeast, as well as the strong negative effects in regions like the Pacific West Coast, Texas, and Florida. Regarding $\beta_2$, there is a weaker effect that operates on a more regional scale, extending from Northwest to South. In addition, machine learning models are able to replicate both the non-linear U-shaped effect and the positive linear effects present in the data-generating processes. Specifically, TabNet, due to its neural network structure, produces smoother decision boundaries which results in smoothed effects both spatially and across feature values. LightGBM and XGBoost, as tree-based models, create less smooth effects due to their discrete decision boundaries. However, they still capture most of the true data-generating process. The Random Forest model performs with lower accuracy and its explanations show weaker spatially varying effects, especially for the process with higher spatial heterogeneity. Global non-linear and linear effects captured in Random Forest are also less accurate compared to the other models. 

\begin{figure}
    \centering
    \includegraphics[width=0.75\linewidth]{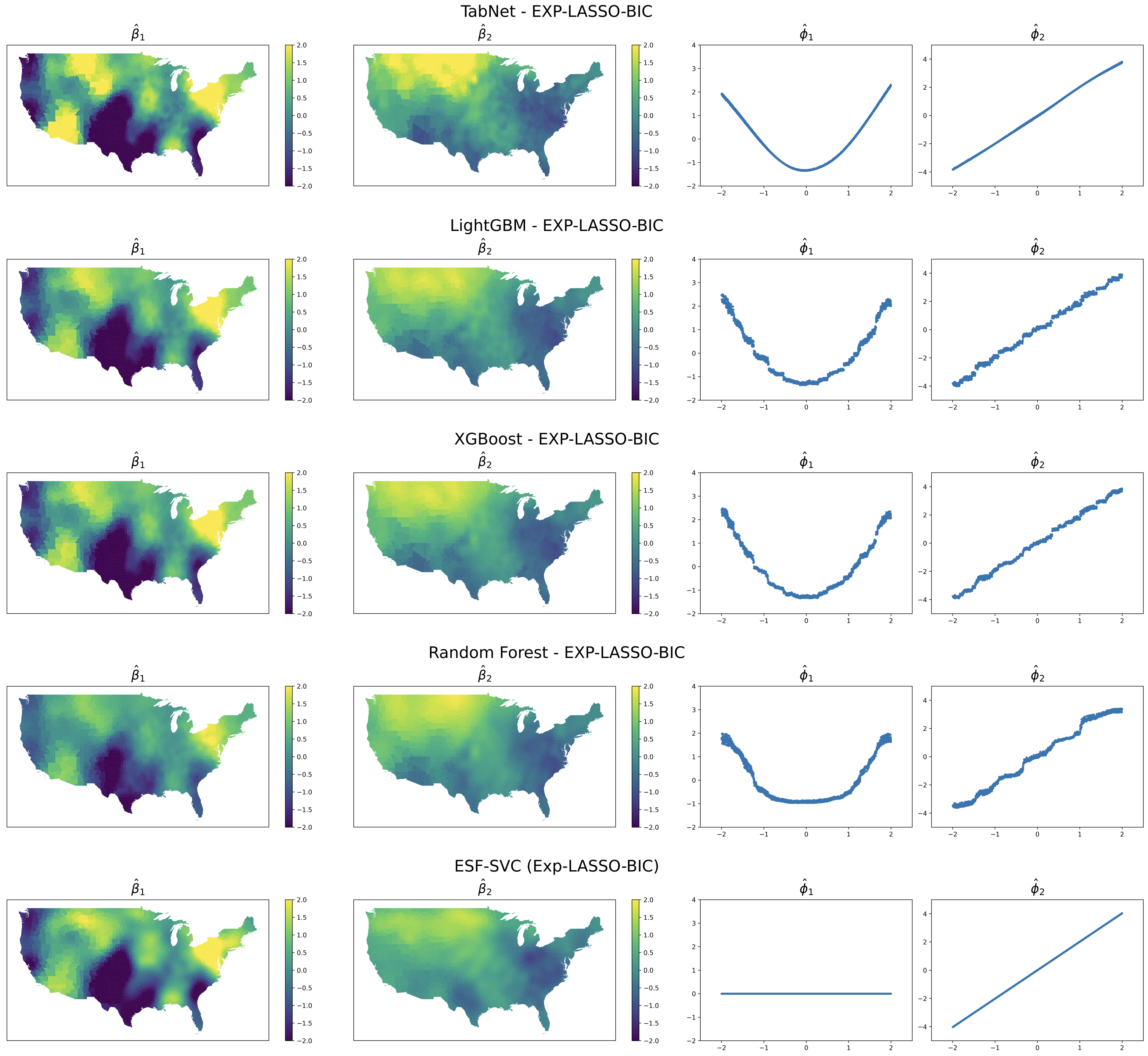}
    \caption{GeoShapley explanations for five models using Moran Eigenvectors (Exp-LASSO-BIC) as input spatial features.}
    \label{fig:geoshapley_us_exp_bic}
\end{figure}

Figure \ref{fig:geoshapley_us_exp_bic}  demonstrates models that use Moran eigenvectors generated from an exponential spatial weights matrix and selected based on the LASSO-BIC criterion (EXP-LASSO-BIC). This approach is chosen because machine learning models with EXP-LASSO-BIC outperform models with other Moran eigenvector settings. The results are generally similar to those obtained using coordinates, though some nuances and inaccuracies are evident. For example, for the explained spatially varying effects, there are artifacts and patterns resulting from Moran Eigenvectors features with small eigenvalues. In the bottom explanations generated by the ESF-SVC model, the lack of non-linearity in the model specification prevents it from capturing non-linear effects. Results for the "Grid" dataset generally mirror those for the "US Counties" dataset, which can be found in Appendix A.

\section{Discussion and Conclusion}

\begin{figure}
    \centering
    \includegraphics[width=0.75\linewidth]{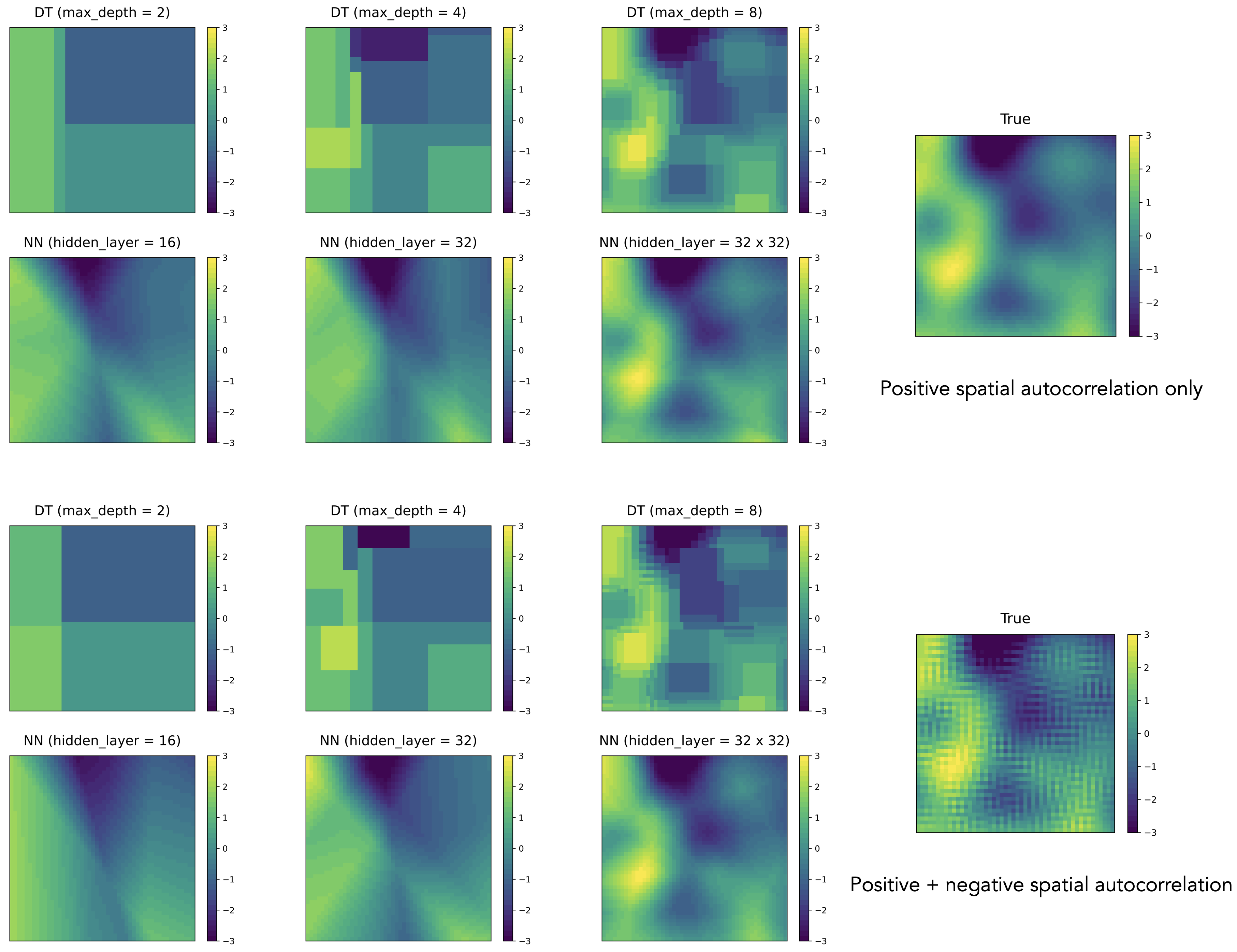}
    \caption{An illustration of using trees and neural networks to approximate spatial patterns from coordinates.}
    \label{fig:dt_nn_appro}
\end{figure}

This paper investigates the effectiveness of using Moran Eigenvectors (ME) as additional features in machine learning models. While existing studies have demonstrated that incorporating Moran Eigenvectors increases model accuracy and effectively captures spatial effects, there is limited comparison of this approach to a simpler alternative: adding coordinates directly into the model. Evidence from experiments based on known data-generating processes and synthetic data shows that adding coordinates alone can easily and effectively account for possible spatially varying effects. One explanation is due to the non-linear nature of machine learning model that transforms coordinates into non-linear space, which further manifests spatial patterns that serve a similar purpose to Moran Eigenvectors. A simple demonstration in Figure \ref{fig:dt_nn_appro} shows that a decision tree model can leverage coordinates to split a spatial surface by increasing the tree's depth. A decision tree with a depth of 8 can approximate the designed spatial pattern quite well. Similarly, the second row of figures in Figure \ref{fig:dt_nn_appro} show that increasing the number of neurons and layers can also replicate spatial patterns with additional details, and the last figure in the second row with 2 hidden layers of 32 neurons largely reflects the spatial pattern. This is due to the fact that both decision tree-based models and neural networks are considered universal approximators (Hornik et al., 1989; Friedman. 2001) that are capable of approximating any functional form with sufficient complexity. Spatial patterns and effects are fundamentally functions of locations and coordinates, therefore it is not too surprising that these machine learning models can approximate spatial effects solely from coordinate information. 

Though Moran Eigenvectors are widely recognized to be useful for accounting for spatial effects in linear models, in machine learning, adding them does not necessarily improve model accuracy and may even make it harder for the model to capture the true processes by introducing a larger number of additional features, which unavoidably makes feature selection more challenging. This is particularly true for TabNet and, to a lesser extent, for tree-based models, as tree-based models are inherently better at feature selection than neural networks (e.g., as discussed in Grinsztajn et al., 2022). On the other hand, using coordinates simply adds only two additional features, which introduces minimal complexity for models to train and converge.

However, it is worth recognizing that Moran Eigenvector approaches are useful in scenarios beyond the experiments presented in this study. First, the experiments are based on spatially varying, nonlinear processes, but other spatial processes exist, such as those govern the generation of network data, including flows of people, vehicles, and goods. Eigenvector spatial filtering can be extended to model network autocorrelation by replacing the spatial weights matrix with a network link matrix. Research has shown that this approach effectively reduces network autocorrelation and improves the accuracy of estimates and inferences in spatial interaction models (Chun, 2008; Chun and Griffith, 2011). Since a network link matrix can differ substantially from a traditional spatial weights matrix (e.g., in a migration network where people move from New York to Florida rather than being constrained by geographic proximity), coordinates alone cannot capture such network autocorrelation. From a machine learning perspective, a graph-based learning framework would be more appropriate for handling this type of data structure (e.g., Zhu et al., 2020; Liang et al., 2023; Luo and Chen, 2024).

Second, although much less common, spatial data can be generated from spatial processes that exhibit negative spatial autocorrelation. This occurs in cases of competitive locational processes such as in economic and ecological activities (Legendre and Fortin, 1989 and Elhorst and Zigova, 2014). More discussions on negative spatial autocorrelation can be found in Griffith 2006 and Griffith 2019. Negative autocorrelation also presents a challenge when using only coordinates in a machine learning model, as the distance between coordinate values does not necessarily indicate process similarity. To illustrate, the bottom half of Figure \ref{fig:dt_nn_appro} shows that when negative spatial autocorrelation is present alongside positive autocorrelation, only the positive component can be adequately captured by coordinates as features, while the negative autocorrelation is largely ignored. If negative autocorrelation is of interest in the studied phenomenon, explicitly incorporating negative Moran eigenvectors would be useful. However, this still poses challenges regarding eigenvector selection in the machine learning setting. Developing better and more efficient methods to account for negative spatial autocorrelation would be an interesting area for future research. 

Regarding model explainability, this paper demonstrates that GeoShapley offers a method to extract parameters from machine learning models, which enables model comparison at the parameter level. This approach facilitates testing models against known data-generating processes—a standard practice in statistics and spatial statistics but is rarely applied in current GeoAI literature. We encourage future work that focuses on developing new GeoAI models or strategies can employ GeoShapley or other explainable AI methods to test against synthetic datasets generated from known data-generating processes before applying them to real-world datasets. 

\textcolor{white}{...}
\textcolor{white}{...}
\textcolor{white}{...}

\section{References}

\setlength{\parindent}{-2em} 
\setlength{\leftskip}{2em} 
\setlength{\parskip}{1em} 

\vspace{\baselineskip}

Anselin, L. (1988). \textit{Spatial Econometrics: Methods and Models}(Vol. 4). Springer Science and Business Media

Anselin, L., and Bera, A. K. (1998). Spatial dependence in linear regression models with an introduction to spatial econometrics. \textit{Statistics textbooks and monographs}, \textit{155}, 237-290.

Arik, S. Ö., and Pfister, T. (2021, May). Tabnet: Attentive interpretable tabular learning. In \textit{Proceedings of the AAAI conference on artificial intelligence} (Vol. 35, No. 8, pp. 6679-6687).
 
Breiman, L. (2001). Random Forests. \textit{Machine Learning}, 45(1), 5-32.

Chen, T., and Guestrin, C. (2016, August). Xgboost: A scalable tree boosting system. In \textit{Proceedings of the 22nd acm sigkdd international conference on knowledge discovery and data mining} (pp. 785-794).

Chun, Y. (2008). Modeling network autocorrelation within migration flows by eigenvector spatial filtering. Journal of Geographical Systems, 10, 317-344.

Chun, Y., and Griffith, D. A. (2011). Modeling network autocorrelation in space–time migration flow data: An eigenvector spatial filtering approach. Annals of the Association of American Geographers, 101(3), 523-536.

Chun, Y. (2014). Analyzing space–time crime incidents using eigenvector spatial filtering: an application to vehicle burglary. \textit{Geographical Analysis}, \textit{46}(2), 165-184.

Diniz‐Filho, J. A. F., and Bini, L. M. (2005). Modelling geographical patterns in species richness using eigenvector‐based spatial filters. \textit{Global Ecology and Biogeography}, \textit{14}(2), 177-185.

Elhorst, J. P., and Zigova, K. (2014). Competition in research activity among economic departments: Evidence by negative spatial autocorrelation. Geographical Analysis, 46(2), 104-125.

Fotheringham, A. S., Brunsdon, C., and Charlton, M. (2002). Geographically Weighted Regression: The Analysis of Spatially Varying Relationships. John Wiley and Sons.

Fotheringham, A. S., Yang, W., and Kang, W. (2017). Multiscale geographically weighted regression (MGWR). Annals of the American Association of Geographers, 107(6), 1247-1265.

Fotheringham, A. S., Oshan, T. M., and Li, Z. (2023). \textit{Multiscale Geographically Weighted Regression: Theory and Practice}. CRC Press.

Friedman, J. H. (2001). Greedy function approximation: a gradient boosting machine. \textit{Annals of statistics}, 1189-1232.

Griffith, D. A. (2006). Hidden negative spatial autocorrelation. Journal of Geographical Systems, 8, 335-355.

Griffith, D. A. (2008). Spatial-filtering-based contributions to a critique of geographically weighted regression (GWR). Environment and Planning A, 40(11), 2751-2769.

Griffith, D., and Chun, Y. (2014). Spatial autocorrelation and spatial filtering. \textit{Handbook of regional science}, 1477-1507.

Griffith, D. A. (2019). Negative spatial autocorrelation: One of the most neglected concepts in spatial statistics. Stats, 2(3), 388-415.

Grinsztajn, L., Oyallon, E., and Varoquaux, G. (2022). Why do tree-based models still outperform deep learning on typical tabular data?. \textit{Advances in neural information processing systems}, \textit{35}, 507-520.

Harris, C., Pymar, R., and Rowat, C. (2021, October). Joint Shapley values: a measure of joint feature importance. In \textit{International Conference on Learning Representations}.

Helbich, M., and Griffith, D. A. (2016). Spatially varying coefficient models in real estate: Eigenvector spatial filtering and alternative approaches. \textit{Computers, Environment and Urban Systems}, \textit{57}, 1-11.

Hengl, T., Heuvelink, G. B., and Rossiter, D. G. (2007). About regression-kriging: From equations to case studies. \textit{Computers and geosciences}, \textit{33}(10), 1301-1315.

Hengl, T., Nussbaum, M., Wright, M. N., Heuvelink, G. B., and Gräler, B. (2018). Random forest as a generic framework for predictive modeling of spatial and spatio-temporal variables. \textit{PeerJ}, \textit{6}, e5518.

Hornik, K., Stinchcombe, M., and White, H. (1989). Multilayer feedforward networks are universal approximators. \textit{Neural networks}, \textit{2}(5), 359-366.

Hu, L., Chun, Y., and Griffith, D. A. (2022). Incorporating spatial autocorrelation into house sale price prediction using random forest model. \textit{Transactions in GIS}, \textit{26}(5), 2123-2144.

Hu, Y., Goodchild, M., Zhu, A. X., Yuan, M., Aydin, O., Bhaduri, B., ... and Newsam, S. (2024). A five-year milestone: reflections on advances and limitations in GeoAI research. \textit{Annals of GIS}, 1-14.

Islam, M. D., Li, B., Lee, C., and Wang, X. (2022). Incorporating spatial information in machine learning: The Moran eigenvector spatial filter approach. \textit{Transactions in GIS}, \textit{26}(2), 902-922.

Kammann, E. E., and Wand, M. P. (2003). Geoadditive models. \textit{Journal of the Royal Statistical Society Series C: Applied Statistics}, \textit{52}(1), 1-18.

Ke, G., Meng, Q., Finley, T., Wang, T., Chen, W., Ma, W., ... and Liu, T. Y. (2017). Lightgbm: A highly efficient gradient boosting decision tree. \textit{Advances in neural information processing systems}, \textit{30}.

Kühn, I., Nobis, M. P., and Durka, W. (2009). Combining spatial and phylogenetic eigenvector filtering in trait analysis. \textit{Global Ecology and Biogeography}, \textit{18}(6), 745-758.

Legendre, P., and Fortin, M. J. (1989). Spatial pattern and ecological analysis. Vegetatio, 80, 107-138.

LeSage, J., and Pace, R. K. (2009). \textit{Introduction to spatial econometrics}. Chapman and Hall/CRC.

Li, B., and Griffith, D. A. (2022). The Moran Spectrum as a Geoinformatic Tupu: implications for the First Law of Geography. \textit{Annals of GIS}, \textit{28}(1), 69-83.

Li, Z. (2022). Extracting spatial effects from machine learning model using local interpretation method: An example of SHAP and XGBoost. \textit{Computers, Environment and Urban Systems}, \textit{96}, 101845.

Li, Z. (2023). Leveraging explainable artificial intelligence and big trip data to understand factors influencing willingness to ridesharing. \textit{Travel Behaviour and Society}, \textit{31}, 284-294.

Li, Z. (2024). GeoShapley: A Game Theory Approach to Measuring Spatial Effects in Machine Learning Models. \textit{Annals of American Association of Geographers.} doi: 10.1080/24694452.2024.2350982.

Liang, Y., Ding, F., Huang, G., and Zhao, Z. (2023). Deep trip generation with graph neural networks for bike sharing system expansion. Transportation Research Part C: Emerging Technologies, 154, 104241.

Liu, X., Kounadi, O., and Zurita-Milla, R. (2022). Incorporating spatial autocorrelation in machine learning models using spatial lag and eigenvector spatial filtering features. \textit{ISPRS International Journal of Geo-Information}, \textit{11}(4), 242.

Luo, M., and Chen, Y. (2024). Simulating inter-city population flows based on graph neural networks. Geocarto International, 39(1), 2331223.

Lundberg, S. M., and Lee, S. I. (2017). A unified approach to interpreting model predictions. \textit{Advances in neural information processing systems}, \textit{30}.

Ma, M., Zhao, G., He, B., Li, Q., Dong, H., Wang, S., and Wang, Z. (2021). XGBoost-based method for flash flood risk assessment. \textit{Journal of Hydrology}, \textit{598}, 126382.

Mai, G., Janowicz, K., Hu, Y., Gao, S., Yan, B., Zhu, R., ... and Lao, N. (2022). A review of location encoding for GeoAI: methods and applications. \textit{International Journal of Geographical Information Science}, \textit{36}(4), 639-673.

Mete, M. O., and Yomralioglu, T. (2023). A Hybrid Approach for Mass Valuation of Residential Properties through Geographic Information Systems and Machine Learning Integration. \textit{Geographical Analysis}, \textit{55}(4), 535-559.

Murakami, D., Yoshida, T., Seya, H., Griffith, D. A., and Yamagata, Y. (2017). A Moran coefficient-based mixed effects approach to investigate spatially varying relationships. \textit{Spatial Statistics}, \textit{19}, 68-89.

Murakami, D. (2017). spmoran: An R package for Moran’s eigenvector-based spatial regression analysis. \textit{arXiv preprint arXiv:1703.04467}.

Oshan, T. M., and Fotheringham, A. S. (2018). A comparison of spatially varying regression coefficient estimates using geographically weighted and spatial‐filter‐based techniques. \textit{Geographical Analysis}, \textit{50}(1), 53-75.

Peng, Z., and Inoue, R. (2022). Identifying Multiple scales of spatial heterogeneity in housing prices based on eigenvector spatial filtering approaches. \textit{ISPRS International Journal of Geo-Information}, \textit{11}(5), 283.

Peng, Z., and Inoue, R. (2024). Multiscale Continuous and Discrete Spatial Heterogeneity Analysis: The Development of a Local Model Combining Eigenvector Spatial Filters and Generalized Lasso Penalties. \textit{Geographical Analysis}, \textit{56}(2), 303-327.

Pradhan, B., Lee, S., Dikshit, A., and Kim, H. (2023). Spatial flood susceptibility mapping using an explainable artificial intelligence (XAI) model. \textit{Geoscience Frontiers}, \textit{14}(6), 101625.

Rey, S. J., Anselin, L., Amaral, P., Arribas‐Bel, D., Cortes, R. X., Gaboardi, J. D., ... and Wolf, L. J. (2022). The PySAL ecosystem: Philosophy and implementation. Geographical Analysis, 54(3), 467-487.

Seya, H., Murakami, D., Tsutsumi, M., and Yamagata, Y. (2015). Application of LASSO to the eigenvector selection problem in eigenvector‐based spatial filtering. \textit{Geographical analysis}, \textit{47}(3), 284-299.

Shi, W., Du, Y., Chang, C. H., Nguyen, S., and Wu, J. (2021). Spatial heterogeneity and economic driving factors of SO2 emissions in China: Evidence from an eigenvector based spatial filtering approach. \textit{Ecological Indicators}, \textit{129}, 108001.

Tan, H., Chen, Y., Wilson, J. P., Zhang, J., Cao, J., and Chu, T. (2020). An eigenvector spatial filtering based spatially varying coefficient model for PM2. 5 concentration estimation: A case study in Yangtze River Delta region of China. \textit{Atmospheric environment}, \textit{223}, 117205.

Tiefelsdorf, M., and Griffith, D. A. (2007). Semiparametric filtering of spatial autocorrelation: the eigenvector approach. \textit{Environment and Planning A}, \textit{39}(5), 1193-1221.

Wang, C., Wu, Q., Weimer, M., and Zhu, E. (2021). Flaml: A fast and lightweight automl library. \textit{Proceedings of Machine Learning and Systems}, \textit{3}, 434-447.

Xue, H., Guo, P., Li, Y., and Ma, J. (2024). Integrating visual factors in crash rate analysis at Intersections: An AutoML and SHAP approach towards cycling safety. \textit{Accident Analysis and Prevention}, \textit{200}, 107544.

Zhong, J., Zhang, X., Gui, K., Wang, Y., Che, H., Shen, X., ... and Zhang, W. (2021). Robust prediction of hourly PM2. 5 from meteorological data using LightGBM. \textit{National science review}, \textit{8}(10), nwaa307.

Zhu, D., Zhang, F., Wang, S., Wang, Y., Cheng, X., Huang, Z., ... and Liu, Y. (2020). Understanding place characteristics in geographic contexts through graph convolutional neural networks. Annals of the American Association of Geographers, 110(2), 408-420.

\appendix
\section{Appendix A }

\begin{figure}[h!]
    \centering
    \includegraphics[width=0.75\linewidth]{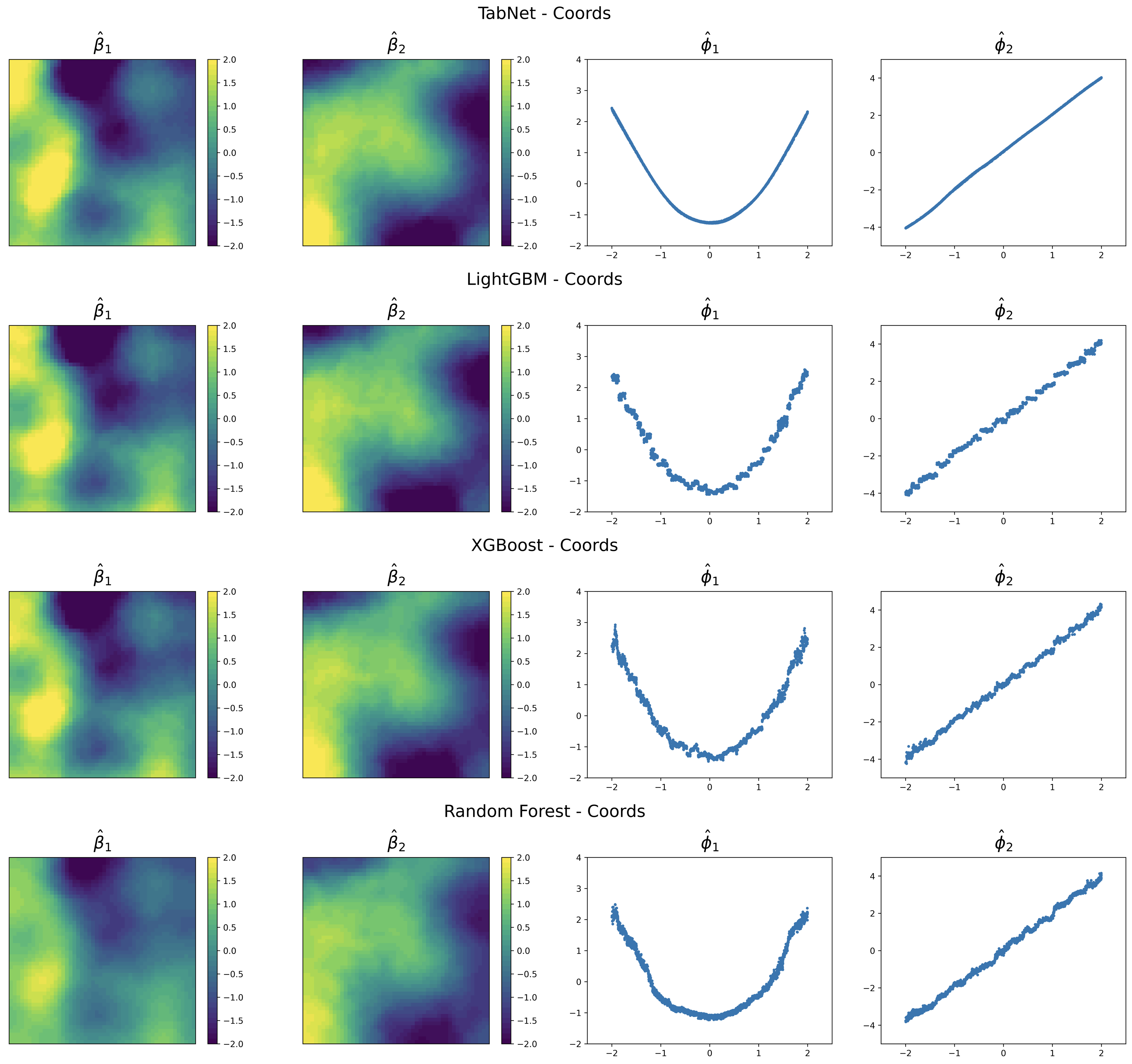}
    \caption{GeoShapley explanations for four models using coordinates as input spatial features.}
    \label{appendix:geoshapley_grid_coords}
\end{figure}

\begin{figure}[h!]
    \centering
    \includegraphics[width=0.75\linewidth]{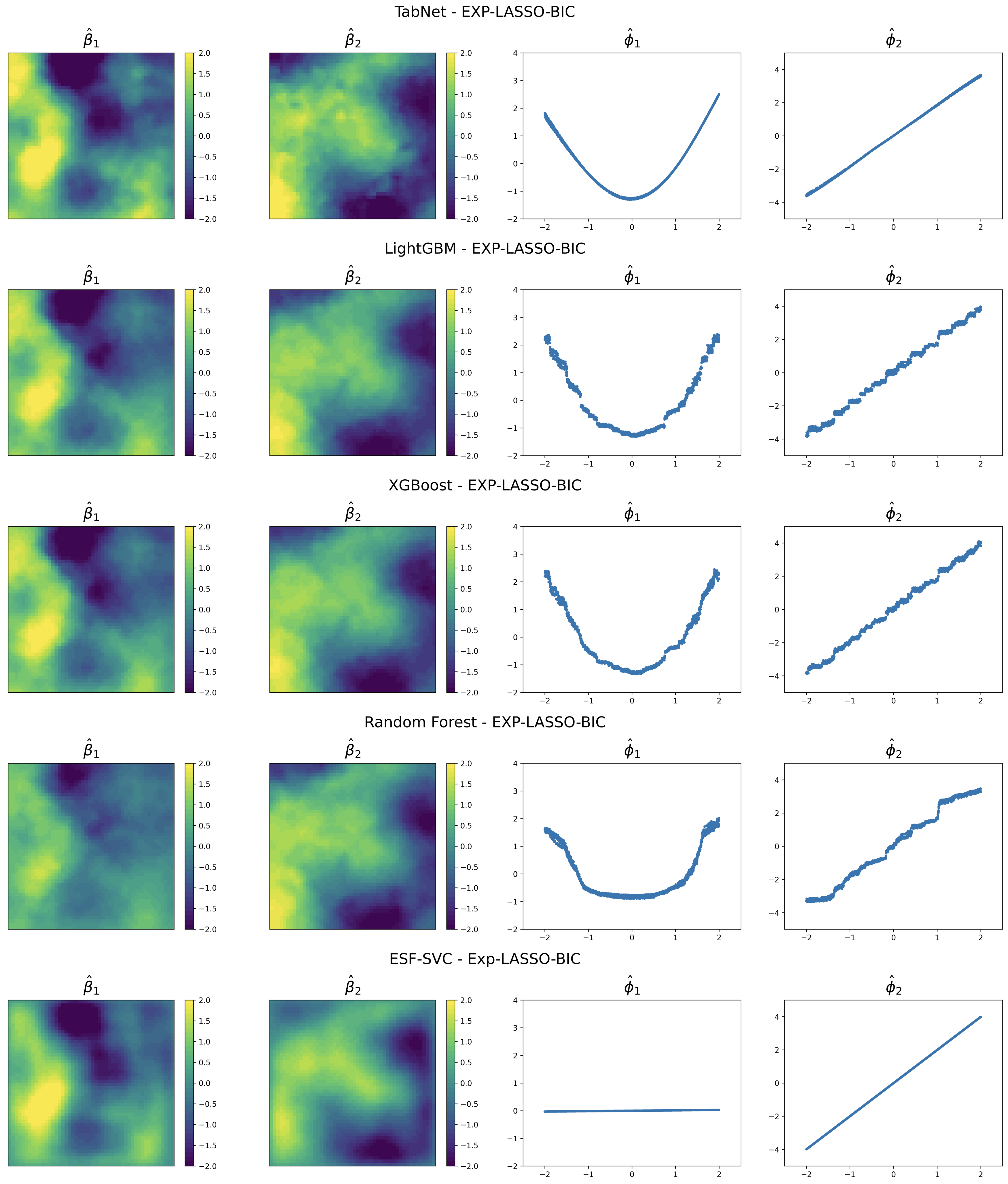}
    \caption{GeoShapley explanations for five models using Moran Eigenvectors (Exp-LASSO-BIC) as input spatial features.}
    \label{appendix:geoshapley_grid_exp_bic}
\end{figure}

\end{document}